\documentclass{article}
\usepackage{spconf,amsmath,graphicx}
\usepackage{multirow}
\usepackage{makecell}
\usepackage{pifont}
\usepackage{amssymb}
\usepackage{xcolor}
\usepackage[export]{adjustbox}

\usepackage{balance}

\title{Neural Network Training Strategy to Enhance Anomaly Detection Performance: A Perspective on Reconstruction Loss Amplification}
\name{
    \begin{tabular}{@{}c@{}}
        YeongHyeon Park$^{1, 2}$, \qquad 
        Sungho Kang$^{1}$ \qquad 
        Myung Jin Kim$^{2}$ \qquad
        Hyeonho Jeong$^{3}$ \\ 
        Hyunkyu Park$^{1}$ \qquad 
        Hyeong Seok Kim$^{2}$ \qquad 
        Juneho Yi$^{1}$\thanks{\hspace*{-0.5cm}This research was supported by SK Planet Co., Ltd. \\Corresponding author: Juneho Yi.}
    \end{tabular}
}
\address{
    $^{1}$Department of Electrical and Computer Engineering, Sungkyunkwan University\\
    $^{2}$SK Planet Co., Ltd.\\
    $^{3}$College of Computing, Sungkyunkwan University\\
    \small{\texttt{\{yeonghyeon, myungjin, beman\}@sk.com, \{sungho369, drake6751, mjss016, jhyi\}@skku.edu}}
}
\begin{document}
%\ninept
%
\maketitle
\begin{abstract}
Unsupervised anomaly detection (UAD) is a widely adopted approach in industry due to rare anomaly occurrences and data imbalance.
A desirable characteristic of an UAD model is contained generalization ability which excels in the reconstruction of seen normal patterns but struggles with unseen anomalies.
Recent studies have pursued to contain the generalization capability of their UAD models in reconstruction from different perspectives, such as design of neural network (NN) structure and training strategy.
In contrast, we note that containing of generalization ability in reconstruction can also be obtained simply from steep-shaped loss landscape.
Motivated by this, we propose a loss landscape sharpening method by amplifying the reconstruction loss, dubbed \textit{Loss AMPlification} (LAMP).
LAMP deforms the loss landscape into a steep shape so the reconstruction error on unseen anomalies becomes greater.
Accordingly, the anomaly detection performance is improved without any change of the NN architecture.
Our findings suggest that LAMP can be easily applied to any reconstruction error metrics in UAD settings where the reconstruction model is trained with anomaly-free samples only.
\end{abstract}
% \textemdash{}i. e., deep learning-based classifier\textemdash{}

\begin{keywords}
Unsupervised anomaly detection, Loss amplification, Loss landscape, Training strategy
\end{keywords}

\section{Introduction}
\label{sec:introduction}

% \begin{figure}[t]
%     \begin{center}
%         \includegraphics*[width=0.95\columnwidth]{figures/proposal_dash22} 
%     \end{center}
%     \caption{Graphical representation of LAMP-applied cases for $\mathcal{L}_{1}$ and $\mathcal{L}_{2}$ loss functions. LAMP makes the model learn constrained generalization boundaries by imposing a strong penalty even if a weak loss, enhancing anomaly detection sensitivity.}
%     \label{fig:loss_curve}
% \end{figure}

\begin{figure}[t]
    \resizebox{\columnwidth}{!}{%
        \setlength\tabcolsep{1.5pt}
        \begin{tabular}{c cc cc}
            \begin{tabular}{c cc cc}
                & \multicolumn{2}{c}{$\mathcal{L}_{2}$} & \multicolumn{2}{c}{$\mathcal{L}_{2}^{LAMP}$} \\
                \vspace*{-0.2cm} \\
                
                \rotatebox[origin=c]{90}{\qquad{}\qquad{}\qquad{}\qquad{}\quad{}Loss landscape} & 
                \multicolumn{2}{c}{\includegraphics*[width=0.5\columnwidth,trim={2.0cm 3.5cm 2.0cm 3.5cm},clip]{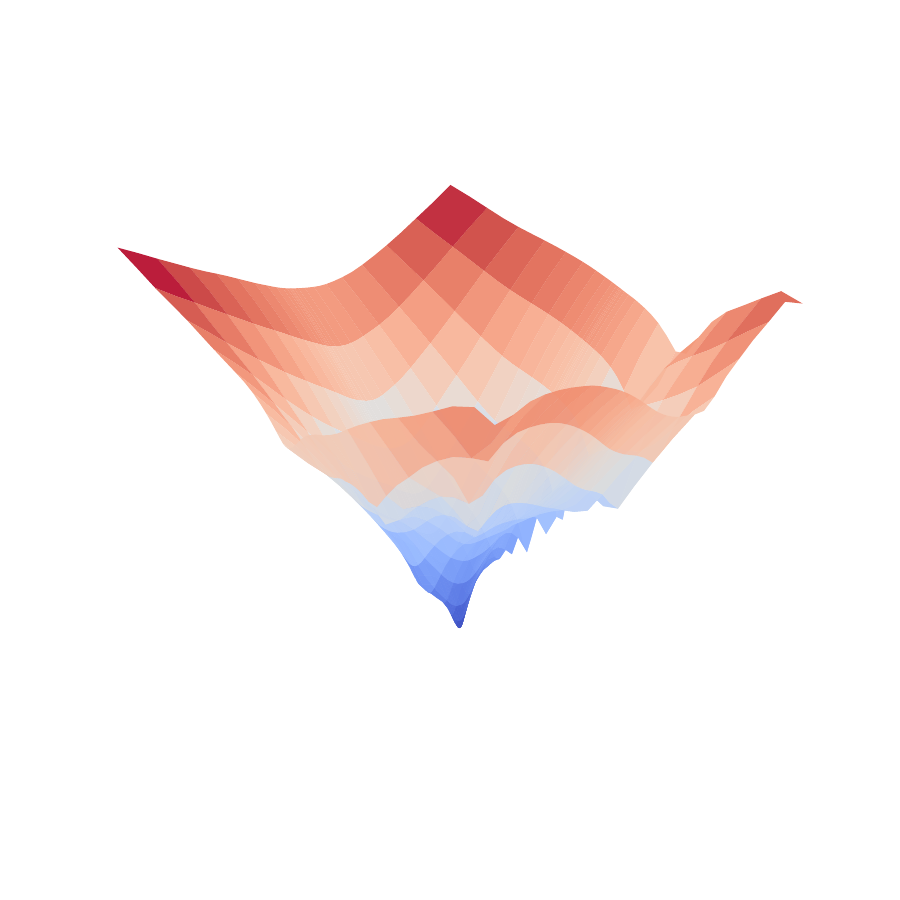}} & 
                \multicolumn{2}{c}{\includegraphics*[width=0.5\columnwidth,trim={2.0cm 3.5cm 2.0cm 3.5cm},clip]{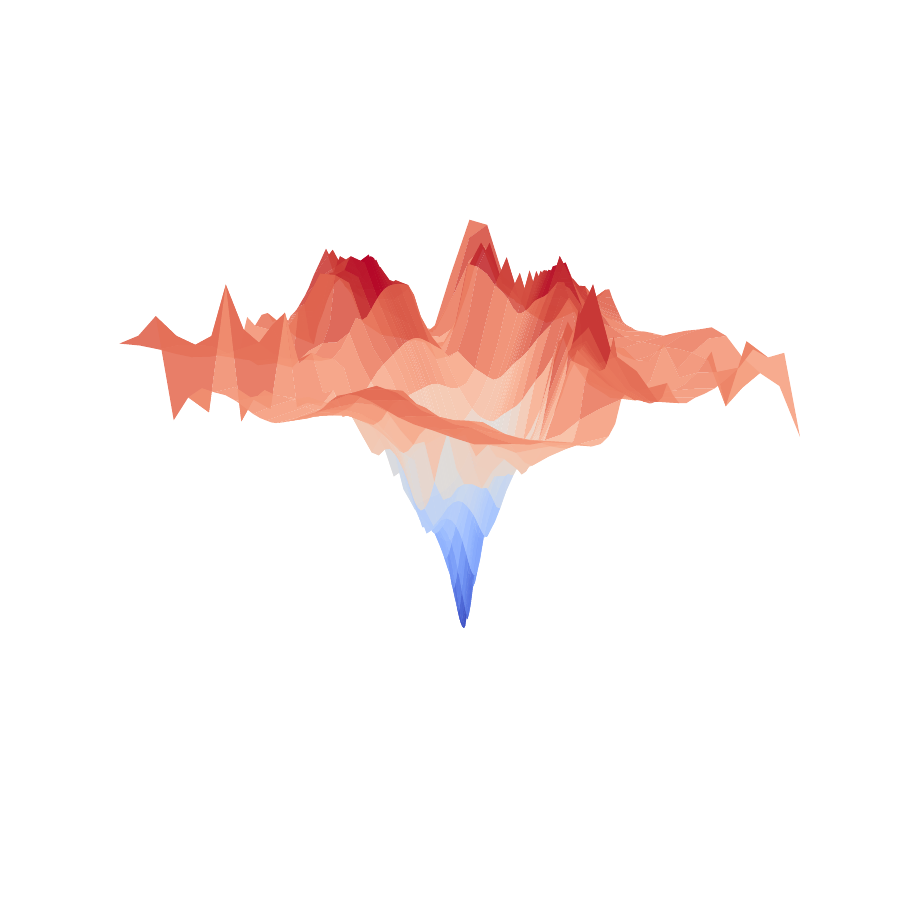}} \\
                \vspace*{-3.0cm} \\
    
                \rotatebox[origin=c]{90}{\qquad{}\qquad{}\qquad{}\qquad{}\qquad{}\qquad{}Contour} & 
                \multicolumn{2}{c}{\includegraphics*[width=0.5\columnwidth,trim={1.5cm 1.3cm 1.5cm 1.3cm},clip]{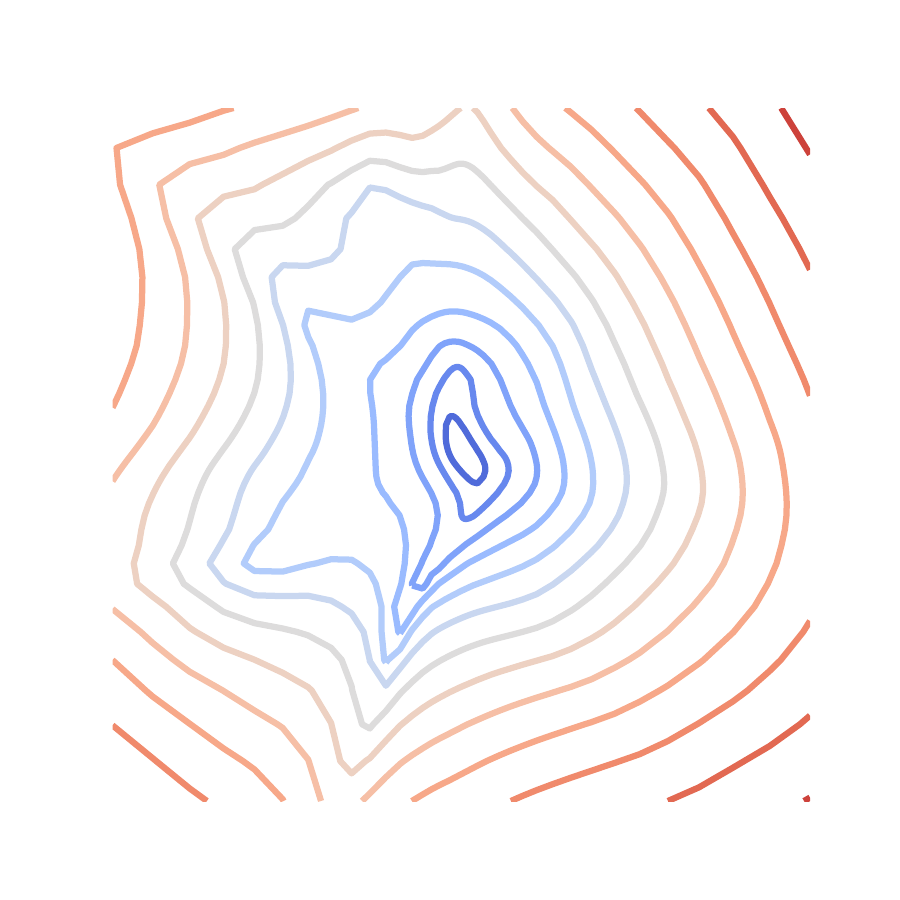}} & 
                \multicolumn{2}{c}{\includegraphics*[width=0.5\columnwidth,trim={1.5cm 1.3cm 1.5cm 1.3cm},clip]{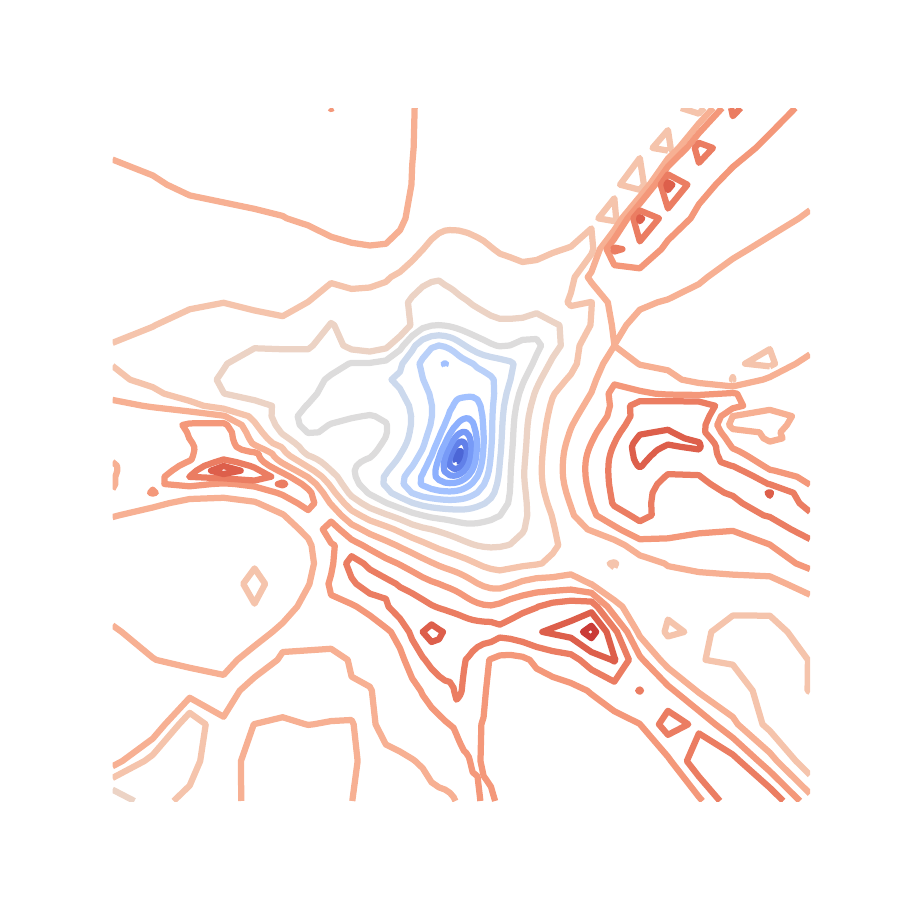}} \\
                \vspace*{-3.0cm} \\
            \end{tabular}
        \end{tabular}
    }
    \caption{Effect of LAMP. The loss landscapes and their contour projections for  $\mathcal{L}_{2}$ and $\mathcal{L}_{2}^{LAMP}$ are shown in the first and second rows, respectively. The loss landscape for an UAD model should be shaped with a steep and sharp form in order to contain the reconstruction generalization ability of the model and enhance the AD performance~\cite{Landscape_Li_NeurIPS18,SWA_Pavel_UAI22,Adversarial_Shadab_BMVC22}.}
    \label{fig:loss_landscape}
    \vspace*{-0.4cm}
\end{figure}

\begin{figure}[t]
    \resizebox{\columnwidth}{!}{%
        \begin{tabular}{c}
        \includegraphics*[width=0.95\columnwidth,trim={0.4cm 0.4cm 0.4cm 0.4cm},clip]{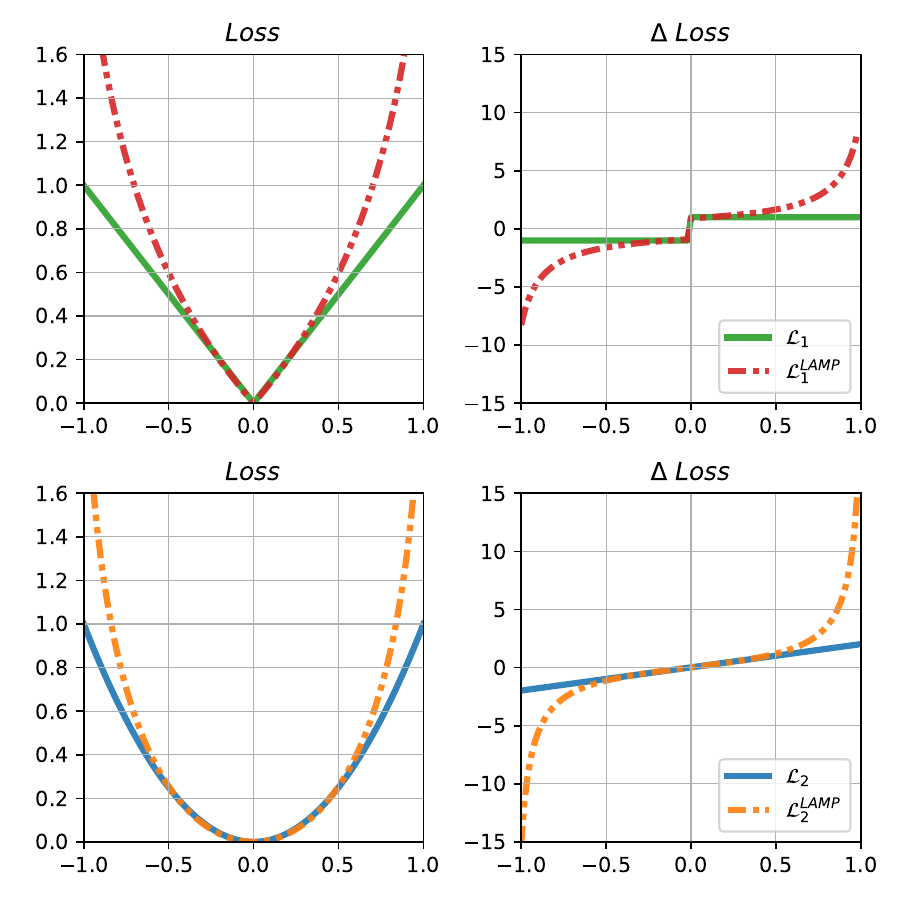} \\
        \vspace*{-0.4cm}
        \end{tabular}
    }
    \caption{Loss curves for LAMP-applied $\mathcal{L}_{1}$ and $\mathcal{L}_{2}$ cases. $\mathcal{L}_{1}$ and $\mathcal{L}_{2}$ cases are shown in the first and second rows, respectively and their gradients are shown in the second column. LAMP imposes a larger penalty than the base loss function, $\mathcal{L}_{base}$.}
    \label{fig:loss_curve}
    \vspace*{-0.4cm}
\end{figure}

Exploiting a reconstruction model trained with anomaly-free samples only is a widely adopted approach to unsupervised anomaly detection (UAD) in various industries due to its capability to resolve the challenges posed by the scarcity of abnormal situations and data imbalance problems.
The desirable characteristic of a trained UAD model is contained generalization ability in reconstruction.
That is, the model should excel in reconstruction of seen normal patterns but struggle with unseen anomalous patterns.

An easy way for containing of the generalization ability of an UAD model in reconstruction is to pour all its reconstruction capability into normal patterns so that there is no room to cover anomalous patterns.
Various methods have been proposed to further improve the anomaly detection (AD) performance by containing generalization ability, but the focus has been mostly on exploring new neural network (NN) structures or extensions.
NN designs in UAD can be divided into three main categories: 
1) generative adversarial networks (GAN) that additionally include a discriminator on a generative model~\cite{GANomaly_Akcay_ACCV18,HPGAN_Park_ETRIJ21,DGM_Tang_MIA21}, 
2) a memory module that can forge normal latent features to reconstruct normal-like images~\cite{MemAE_Gong_ICCV19,BWMem_Hou_ICCV21,Mem_Kim_ICASSP23,SQUID_Xiang_CVPR23}, and 
3) online knowledge distillation to prevent the model from generating a fixed, constant normal image regardless of changing input~\cite{RevDistill_Deng_CVPR22,SQUID_Xiang_CVPR23,MOKD_Song_CVPR23}.

In contrast, we exploit the research results reported in~\cite{Landscape_Li_NeurIPS18} that the generalization ability of NNs is related with the shape of a loss landscape.
They report that in a classification task, NNs with smooth-shaped loss landscape show better generalization ability compared to sharp shapes.
Their loss landscape visualization method is also utilized in generative models~\cite{Adversarial_Shadab_BMVC22} and can also be applied to our reconstruction model.
The loss landscape for an UAD model should be shaped with a steep and sharp form in order to contain the reconstruction generalization ability of the model and enhance the AD performance.

Based on the observation that reconstruction loss amplification causes a sharp shape of the loss landscapes, we propose a method of only changing the reconstruction loss function via amplification, dubbed \textit{Loss AMPlification} (LAMP).
When LAMP is applied, it actually transforms the loss landscape to a sharp form as shown in Fig.~\ref{fig:loss_landscape}.
To verify the legitimacy of our method, we compare the loss landscape using the same encoder/decoder but with different loss functions and batch sizes.
For the comparison, the MNIST dataset~\cite{MNIST_Yan_PIEEE98} is used.
We confirm that when LAMP is applied to the reconstruction error metric for training, it not only transforms the loss landscape into a steeper shape for all batch sizes but also actually enhances the AD performance.
We conduct additional experiments for the MVTec AD dataset~\cite{MVTecAD_Paul_CVPR19} dealing with 15 AD tasks considering combinations of different loss functions and optimizers.
The experimental results show that the AD performance is improved in most cases.

Extensive experiments demonstrate that the application of LAMP leads to an improved AD performance, which is achieved via only loss amplification without the structural change or expansion of NNs. 
LAMP can be easily and safely applied across any reconstruction error metrics when training NNs in UAD settings.

\section{Proposed Method}
\label{sec:propose}

\subsection{Reconstruction generalization for UAD}
\label{subsec:uad}

Loss landscape visualization can provide insight to relate the shape of the loss landscape to the reconstruction generalization ability of an NN model.
When the loss landscape is smooth, a reconstruction model has high generalization ability. 
High generalization ability means that unseen patterns can be well reconstructed at the test time.

However, in UAD, contained generalization ability is crucial because the criterion for determining whether an input sample is defective relies on the magnitude of the reconstruction error.
Note that, when an UAD model has high generalization ability, it will cause the model to reconstruct unseen patterns accurately and fail to detect defective products due to the small reconstruction error.

In this paper, we propose reconstruction loss amplification as a simple way to affect the generalization ability of an UAD model in reconstruction without altering the structure of the NNs or training strategy.

\subsection{Loss amplification}
\label{subsec:lamp}

The proposed method, LAMP, is a simple trick that can improve the AD performance by just amplifying the base loss function $\mathcal{L}_{base}$.
Note that, $\mathcal{L}_{2}$, $\mathcal{L}_{1}$, $\mathcal{L}_{SSIM}$, etc. can be adopted as $\mathcal{L}_{base}$.
LAMP is formulated in \eqref{eq:lamp_loss}. 
Instead of increasing the learning rate or weighting each loss term, LAMP imposes a larger penalty than the base loss function, $\mathcal{L}_{base}$.

\begin{equation}
    \begin{aligned}
        \mathcal{L}_{base}^{LAMP}(y, \hat{y}) = \sum_{h=1}^{H}\sum_{w=1}^{W}\sum_{c=1}^{C}-\log\Bigl(1 - \mathcal{L}_{base}(y, \hat{y})\Bigr), \\
        w.r.t.\ \ y \in \mathbb{R}^{H\times{}W\times{}C}
    \end{aligned}
    \label{eq:lamp_loss}
\end{equation}

LAMP makes gradients steeper than the base loss function as shown in Fig.~\ref{fig:loss_curve}, accelerating loss convergence.
This steeper gradient transforms the loss landscape shape of an UAD model into a sharp form, containing the reconstruction generalization ability.
Note that we can safely amplify the reconstruction loss because an UAD model is only trained using anomaly-free samples. 

\begin{figure}[t]
    \resizebox{\columnwidth}{!}{%
        \setlength\tabcolsep{0.3pt}
        \begin{tabular}{cc}
            \begin{tabular}{c}
                \rotatebox[origin=c]{90}{$\mathcal{L}_{2}$}
            \end{tabular} &
            \begin{tabular}{ccc}
                BS=128 & BS=16 & BS=4 \\
                \vspace*{-0.3cm} \\
                \includegraphics*[width=0.35\columnwidth,trim={1.5cm 1.3cm 1.5cm 1.3cm},clip]{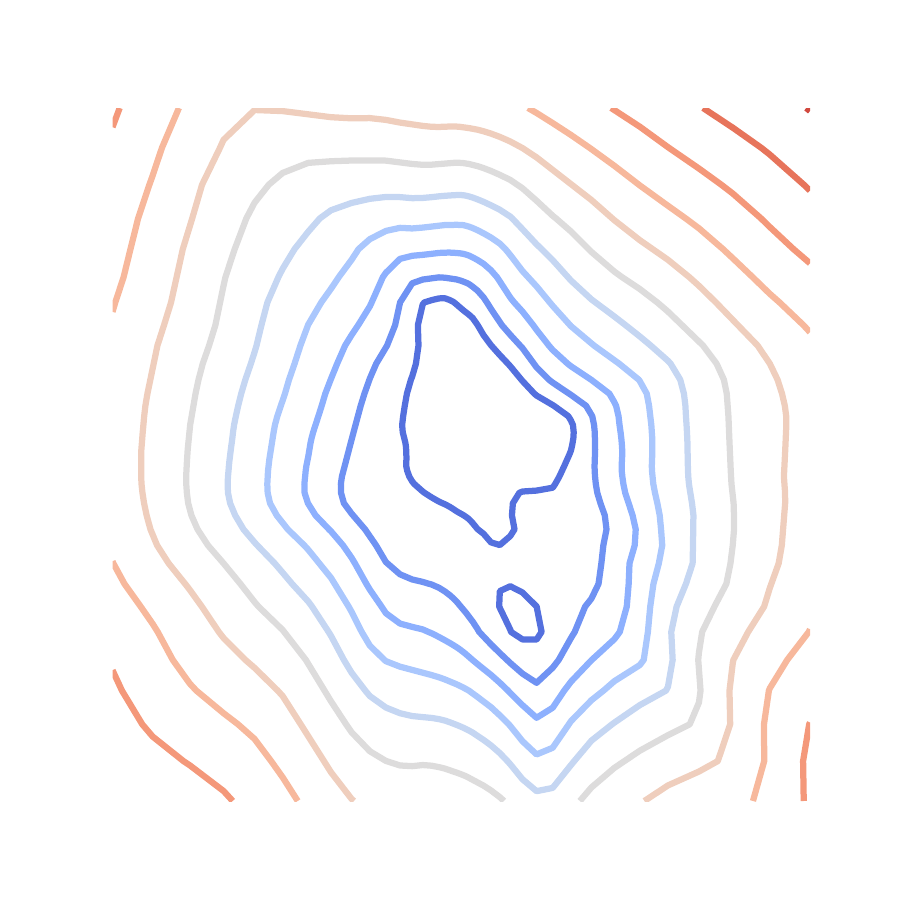} &
                \includegraphics*[width=0.35\columnwidth,trim={1.5cm 1.3cm 1.5cm 1.3cm},clip]{figures/landscape/bs0016_lr5em2_con_l2.pdf} &
                \includegraphics*[width=0.35\columnwidth,trim={1.5cm 1.3cm 1.5cm 1.3cm},clip]{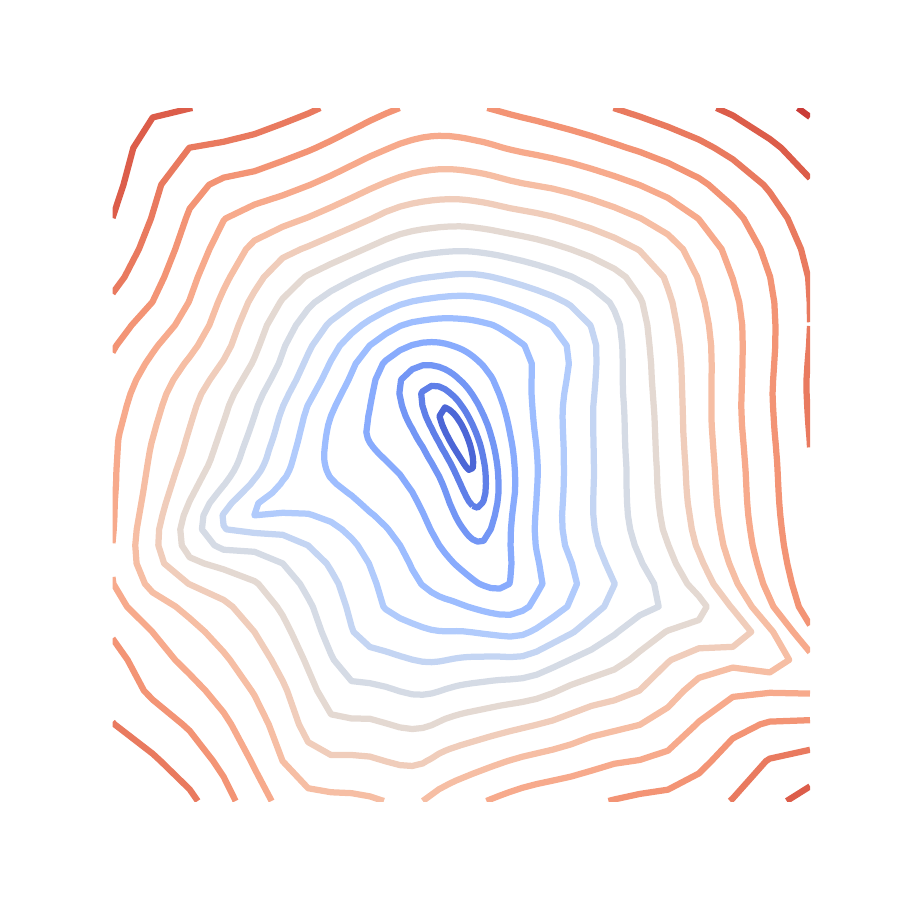} \\
            \end{tabular} \\
            \begin{tabular}{c}
                \rotatebox[origin=c]{90}{$\mathcal{L}_{2}^{LAMP}$}
            \end{tabular} &
            \begin{tabular}{ccc} % ,cfbox=red 0.75pt 0pt
                \includegraphics*[width=0.35\columnwidth,trim={1.5cm 1.3cm 1.5cm 1.3cm},clip]{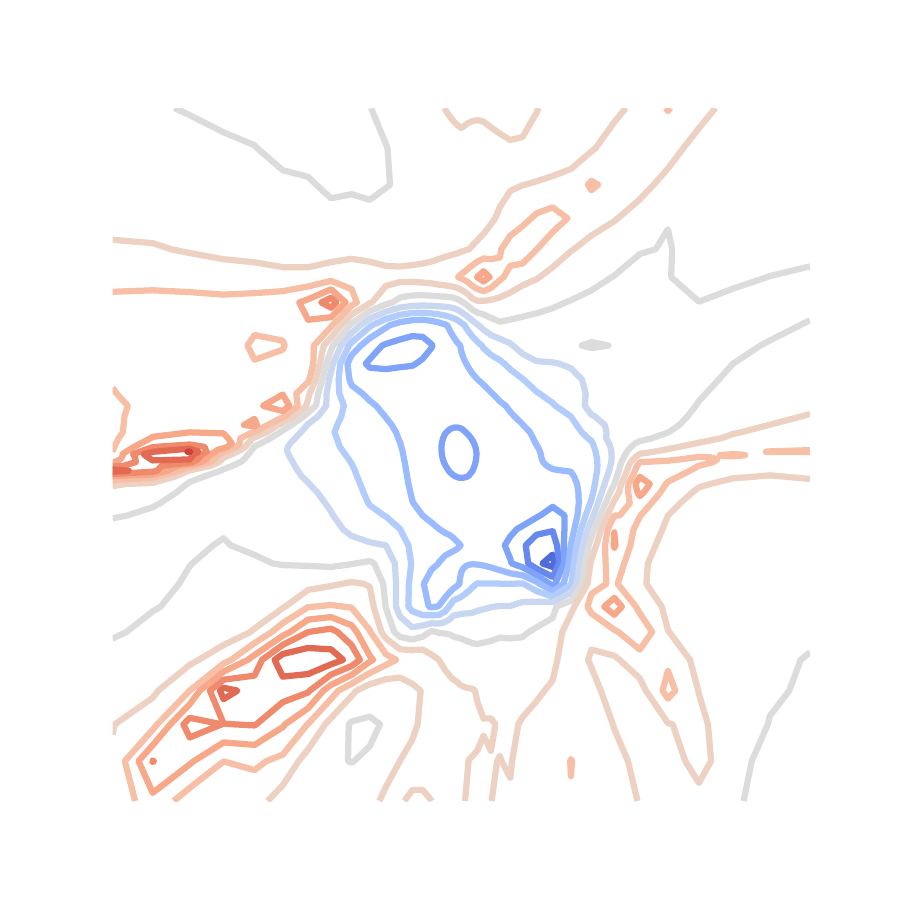} &
                \includegraphics*[width=0.35\columnwidth,trim={1.5cm 1.3cm 1.5cm 1.3cm},clip]{figures/landscape/bs0016_lr5em2_con_fll2.pdf} &
                \includegraphics*[width=0.35\columnwidth,trim={1.5cm 1.3cm 1.5cm 1.3cm},clip]{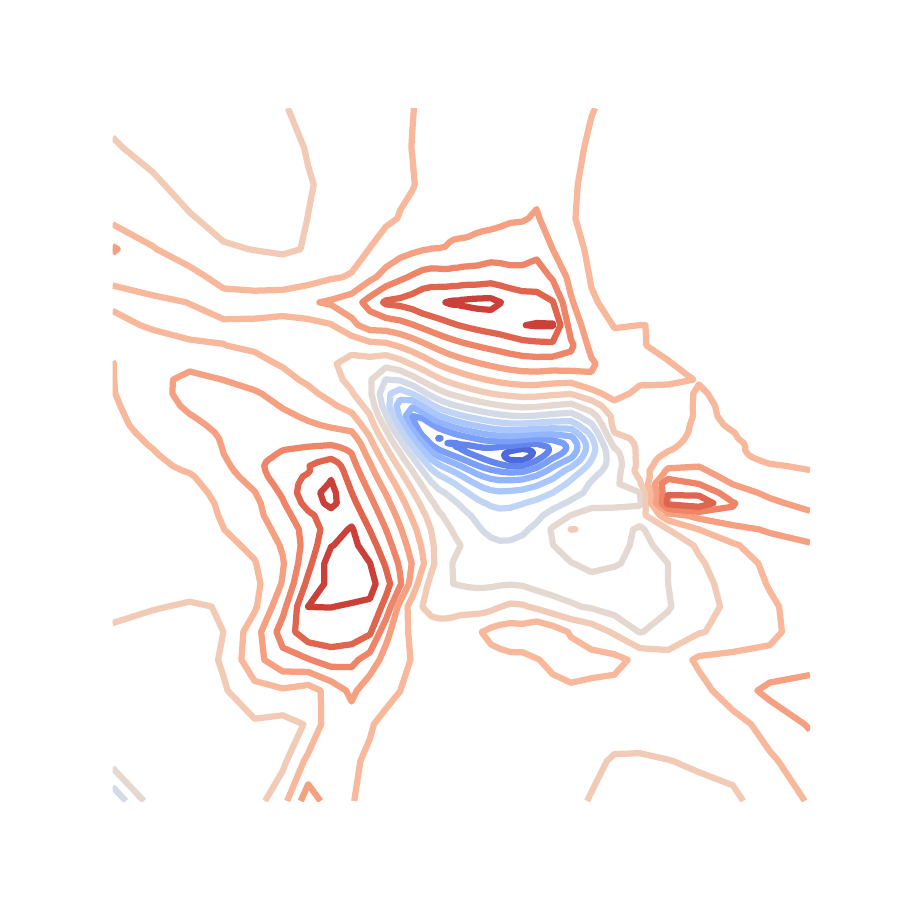} \\
            \end{tabular}
            \vspace*{-0.2cm} \\
        \end{tabular}
    }
    \caption{The contour representation of the loss landscapes with three batch size (BS) conditions for $\mathcal{L}_{2}$ and $\mathcal{L}_{2}^{LAMP}$. It is known that the generalization ability is contained when the BS is small~\cite{Landscape_Li_NeurIPS18}. We experimentally confirm the same effect for LAMP.}
    % A landscape that is expected to perform best is marked with a red box.
    \label{fig:loss_contour}
    \vspace*{-0.4cm}
\end{figure}

\subsection{Scaling trick}
\label{subsec:scaling}

The input of the log function in LAMP must be guaranteed to be positive.
For this, we should ensure that a negative value does not occur for `$1 - \mathcal{L}_{base}$' operation.
We use a simple scaling trick that normalizes $\mathcal{L}_{base}$ between 0 and 1 as in \eqref{eq:scale_trick}.
We also multiply the coefficients `$1-\epsilon$' to adjust $\mathcal{L}_{base}^{\prime}$ to slightly less than 1 which prevents `$\log(1-\mathcal{L}_{base}^{\prime})$' from exploding.

\begin{equation}
    \begin{aligned}
        \mathcal{L}_{base}^{\prime}(y, \hat{y}) = \frac{\mathcal{L}_{base}(y, \hat{y})}{\max{(\mathcal{L}_{base}(y, \hat{y}))}}(1-\epsilon)%, \ \ y \in \mathbb{R}^{H\times{}W\times{}C}
    \end{aligned}
    \label{eq:scale_trick}
\end{equation}

% \vspace*{0.5cm}
\section{Experiments}
\label{sec:experiments}

\subsection{Experimental setup}
\label{subsec:ladscape} 
A basic experiment is designed with the MNIST dataset~\cite{MNIST_Yan_PIEEE98} to see whether LAMP improves the AD performance.
The MNIST dataset~\cite{MNIST_Yan_PIEEE98} is originally provided for classification of digits into ten classes, but we redesign it for AD experiments by setting one class as normal and the other nine as abnormal. 
For example, if the class `0' is set as normal, the AE will be trained by `0' only with the intent of filtering out unseen other nine digit classes `1-9' by large reconstruction error. 
We also report experimental results using the industrial dataset, MVTec AD~\cite{MVTecAD_Paul_CVPR19}.
The training set only includes anomaly-free samples, but the test set includes both anomaly-free and anomalous samples.

% \noindent \textbf{Implementation details.}
\textbf{Implementation details.}
We design a simple convolutional autoencoder (AE) referring to the previous study~\cite{AE_Collin_ICPR20} while discarding skip connections which enables better generalization ability.
Basically, the AE is structured with six layers for encoder and decoder respectively, but for low-resolution datasets such as MNIST~\cite{MNIST_Yan_PIEEE98}, we change it into a four-layered structure for each.
Note that we repeat `convolution $\rightarrow$ batch normalization $\rightarrow$ leaky ReLU activation' for the encoder and `upsampling $\rightarrow$ convolution $\rightarrow$ batch normalization $\rightarrow$ leaky ReLU activation' for the decoder.

% \noindent \textbf{Training conditions.}
\textbf{Training conditions.}
In all AD experiments, we perform hyperparameter tuning to compare the best performance of each model. 
The following hyperparameters are tuned: 1) batch size, 2) a number of patches for patch-wise reconstruction~\cite{PatchRecon_LEE_Access21}, 3) learning rate, and 4) kernel size.
For the experiment using the industrial dataset, MVTec AD~\cite{MVTecAD_Paul_CVPR19}, we use additional training conditions: 1) base loss functions ($\mathcal{L}_{2}$, $\mathcal{L}_{1}$, and $\mathcal{L}_{SSIM}$), and 2) NN optimizers (SGD~\cite{SGD_Bottou_NeurIPS07}, RMSprop~\cite{RMS_Hinton_Coursera12}, and Adam~\cite{Adam_Kingma_arXiv14}).

% \noindent \textbf{Evaluation metric.}
\textbf{Evaluation metric.}
We use the area under the receiver operating characteristic curve (AUROC)~\cite{AUROC_Tom_PRL06} as an evaluation metric in the AD experiments.
The reconstruction error is also called an anomaly score in the AD task.
In this study, $\mathcal{L}_{2}$ between the input $y$ and reconstruction output $\hat{y}$ is used as an anomaly score.
AUROC will be close to 1 when the reconstruction errors of the AE for the unseen anomalous patterns are relatively larger compared to errors in normal pattern reconstruction.

\begin{table}[t]
    \centering
    % \scriptsize
    \caption{The average performance for ten AD tasks using the MNIST dataset~\cite{MNIST_Yan_PIEEE98}. LAMP-applied loss function, $\mathcal{L}_{2}^{LAMP}$, always outperforms $\mathcal{L}_{2}$.}
    \vspace*{0.3cm}
    \resizebox{\columnwidth}{!}{%
        \begin{tabular}{l|cccccc}
            \hline
                \multirow{2}{*}{\textbf{Loss}} & \multicolumn{6}{c}{\textbf{Batch size}} \\
            \cline{2-7}
                 &  \textbf{1024} & \textbf{128} &  \textbf{32} & \textbf{16} &  \textbf{4} & \textbf{1} \\
            \hline
                $\mathcal{L}_{2}$ & 0.658 & 0.919 & 0.921 & 0.926 & 0.931 & 0.919 \\ 
                \vspace*{-0.15cm} \\
                $\mathcal{L}_{2}^{LAMP}$ & \textbf{0.712} & \textbf{0.925} & \textbf{0.929} & \textbf{0.929} & \textbf{0.932} & \textbf{0.927} \\ 
            \hline
        \end{tabular}
    }
    \label{table:mnist_batch}
    \vspace*{-0.4cm}
\end{table}

\begin{table*}[t]
    \centering
    \caption{Summary of the maximum AUROC with hyperparameters tuned for the MVTec AD dataset~\cite{MVTecAD_Paul_CVPR19}. The average AD performance is equal or greater when LAMP is applied in 5 out of 9 experimental settings for three base loss functions ($\mathcal{L}_{2}$, $\mathcal{L}_{1}$, and $\mathcal{L}_{SSIM}$) and three optimizers (SGD~\cite{SGD_Bottou_NeurIPS07}, RMSprop~\cite{RMS_Hinton_Coursera12}, and Adam~\cite{Adam_Kingma_arXiv14}).}
    \vspace*{0.2cm}
    \setlength{\tabcolsep}{4pt}
    \resizebox{\textwidth}{!}{%
        \begin{tabular}{l||c|c|c|c|c|c|c|c|c||c}
            \hline 
                \textbf{Training} & \multicolumn{3}{c|}{$\mathcal{L}_{2}$ $\rightarrow{}$ $\mathcal{L}_{2}^{LAMP}$} & \multicolumn{3}{c|}{$\mathcal{L}_{1}$ $\rightarrow{}$ $\mathcal{L}_{1}^{LAMP}$} & \multicolumn{3}{c||}{$\mathcal{L}_{SSIM}$~\cite{Adam_Kingma_arXiv14} $\rightarrow{}$ $\mathcal{L}_{SSIM}^{LAMP}$} & \textbf{Best} \\
                \cline{1-10}
                \textbf{Optimizer} & \textbf{SGD} & \textbf{RMSprop} & \textbf{Adam} & \textbf{SGD} & \textbf{RMSprop} & \textbf{Adam} & \textbf{SGD} & \textbf{RMSprop} & \textbf{Adam} & $\mathcal{L}_{base}$ $\rightarrow{}$ $\mathcal{L}_{base}^{LAMP}$ \\
            \hline
            \hline
                Bottle  & \underline{0.987} $\rightarrow{}$ 0.983 & \underline{0.990} $\rightarrow{}$ \underline{0.990} & \underline{0.993} $\rightarrow{}$ 0.991 & 0.989 $\rightarrow{}$ \underline{0.992} & 0.993 $\rightarrow{}$ \underline{0.994} & \underline{0.994} $\rightarrow{}$ 0.992 & \underline{0.983} $\rightarrow{}$ 0.980 & \underline{0.994} $\rightarrow{}$ \underline{0.994} & \underline{0.994} $\rightarrow{}$ 0.993 &  \textbf{0.994} $\rightarrow{}$ \textbf{0.994} \\
                Cable  & 0.806 $\rightarrow{}$ \underline{0.813} & \underline{0.832} $\rightarrow{}$ 0.830 & \underline{0.817} $\rightarrow{}$ 0.812 & \underline{0.823} $\rightarrow{}$ 0.790 & \underline{0.832} $\rightarrow{}$ 0.830 & \underline{0.835} $\rightarrow{}$ 0.823 & 0.728 $\rightarrow{}$ \underline{0.755} & \underline{0.798} $\rightarrow{}$ 0.781 & \underline{0.811} $\rightarrow{}$ 0.792 &  \textbf{0.835} $\rightarrow{}$ 0.830 \\
                Capsule  & \underline{0.816} $\rightarrow{}$ 0.791 & 0.782 $\rightarrow{}$ \underline{0.800} & \underline{0.810} $\rightarrow{}$ 0.775 & 0.764 $\rightarrow{}$ \underline{0.799} & 0.757 $\rightarrow{}$ \underline{0.811} & 0.801 $\rightarrow{}$ \underline{0.816} & \underline{0.801} $\rightarrow{}$ \underline{0.801} & \underline{0.798} $\rightarrow{}$ 0.786 & 0.793 $\rightarrow{}$ \underline{0.825} &  0.816 $\rightarrow{}$ \textbf{0.825} \\
                Hazelnut  & 0.980 $\rightarrow{}$ \underline{0.981} & 0.974 $\rightarrow{}$ \underline{0.993} & 0.965 $\rightarrow{}$ \underline{0.974} & \underline{0.982} $\rightarrow{}$ 0.981 & 0.984 $\rightarrow{}$ \underline{0.988} & 0.972 $\rightarrow{}$ \underline{0.983} & 0.894 $\rightarrow{}$ \underline{0.938} & 0.956 $\rightarrow{}$ \underline{0.959} & 0.947 $\rightarrow{}$ \underline{0.951} &  0.984 $\rightarrow{}$ \textbf{0.993} \\
                Metal nut  & 0.637 $\rightarrow{}$ \underline{0.665} & \underline{0.762} $\rightarrow{}$ 0.691 & \underline{0.785} $\rightarrow{}$ 0.694 & \underline{0.711} $\rightarrow{}$ 0.684 & \underline{0.685} $\rightarrow{}$ 0.677 & \underline{0.718} $\rightarrow{}$ 0.708 & \underline{0.728} $\rightarrow{}$ 0.709 & 0.776 $\rightarrow{}$ \underline{0.782} & 0.715 $\rightarrow{}$ \underline{0.819} &  0.785 $\rightarrow{}$ \textbf{0.819} \\
                Pill  & \underline{0.810} $\rightarrow{}$ 0.803 & \underline{0.864} $\rightarrow{}$ \underline{0.864} & 0.860 $\rightarrow{}$ \underline{0.885} & \underline{0.856} $\rightarrow{}$ 0.845 & 0.867 $\rightarrow{}$ \underline{0.874} & 0.834 $\rightarrow{}$ \underline{0.836} & 0.824 $\rightarrow{}$ \underline{0.827} & \underline{0.857} $\rightarrow{}$ 0.832 & \underline{0.837} $\rightarrow{}$ 0.830 &  0.867 $\rightarrow{}$ \textbf{0.885} \\
                Screw  & 0.817 $\rightarrow{}$ \underline{0.827} & \underline{0.826} $\rightarrow{}$ \underline{0.826} & \underline{0.831} $\rightarrow{}$ 0.804 & 0.774 $\rightarrow{}$ \underline{0.827} & \underline{0.826} $\rightarrow{}$ \underline{0.826} & 0.724 $\rightarrow{}$ \underline{0.831} & \underline{0.752} $\rightarrow{}$ 0.712 & 0.827 $\rightarrow{}$ \underline{0.832} & \underline{0.789} $\rightarrow{}$ 0.788 &  0.831 $\rightarrow{}$ \textbf{0.832} \\
                Toothbrush  & \underline{0.969} $\rightarrow{}$ 0.950 & 0.956 $\rightarrow{}$ \underline{0.969} & \underline{0.981} $\rightarrow{}$ 0.978 & 0.956 $\rightarrow{}$ \underline{0.964} & 0.919 $\rightarrow{}$ \underline{0.964} & 0.983 $\rightarrow{}$ \underline{0.986} & \underline{0.850} $\rightarrow{}$ 0.844 & 0.958 $\rightarrow{}$ \underline{0.972} & \underline{0.972} $\rightarrow{}$ 0.958 &  0.983 $\rightarrow{}$ \textbf{0.986} \\
                Transistor  & 0.866 $\rightarrow{}$ \underline{0.885} & 0.889 $\rightarrow{}$ \underline{0.901} & 0.906 $\rightarrow{}$ \underline{0.932} & 0.882 $\rightarrow{}$ \underline{0.899} & \underline{0.894} $\rightarrow{}$ 0.881 & \underline{0.902} $\rightarrow{}$ \underline{0.902} & 0.825 $\rightarrow{}$ \underline{0.847} & 0.879 $\rightarrow{}$ \underline{0.888} & \underline{0.895} $\rightarrow{}$ 0.888 &  0.906 $\rightarrow{}$ \textbf{0.932} \\
                Zipper  & 0.860 $\rightarrow{}$ \underline{0.893} & 0.864 $\rightarrow{}$ \underline{0.867} & \underline{0.918} $\rightarrow{}$ 0.859 & 0.876 $\rightarrow{}$ \underline{0.887} & 0.839 $\rightarrow{}$ \underline{0.855} & \underline{0.914} $\rightarrow{}$ 0.907 & \underline{0.829} $\rightarrow{}$ 0.809 & \underline{0.924} $\rightarrow{}$ 0.923 & 0.929 $\rightarrow{}$ \underline{0.938} &  0.929 $\rightarrow{}$ \textbf{0.938} \\
            \hline
                Carpet  & 0.709 $\rightarrow{}$ \underline{0.721} & \underline{0.872} $\rightarrow{}$ 0.856 & \underline{0.677} $\rightarrow{}$ 0.657 & 0.640 $\rightarrow{}$ \underline{0.702} & \underline{0.921} $\rightarrow{}$ 0.806 & 0.652 $\rightarrow{}$ \underline{0.671} & 0.654 $\rightarrow{}$ \underline{0.669} & 0.610 $\rightarrow{}$ \underline{0.621} & \underline{0.643} $\rightarrow{}$ 0.641 &  \textbf{0.921} $\rightarrow{}$ 0.856 \\
                Grid  & \underline{0.791} $\rightarrow{}$ 0.787 & 0.868 $\rightarrow{}$ \underline{0.888} & \underline{0.920} $\rightarrow{}$ 0.894 & \underline{0.758} $\rightarrow{}$ 0.722 & 0.859 $\rightarrow{}$ \underline{0.868} & 0.869 $\rightarrow{}$ \underline{0.904} & \underline{0.652} $\rightarrow{}$ 0.651 & \underline{0.895} $\rightarrow{}$ 0.825 & \underline{0.880} $\rightarrow{}$ 0.833 &  \textbf{0.920} $\rightarrow{}$ 0.904 \\
                Leather  & \underline{0.988} $\rightarrow{}$ 0.983 & 0.967 $\rightarrow{}$ \underline{0.978} & \underline{0.997} $\rightarrow{}$ 0.993 & \underline{0.986} $\rightarrow{}$ 0.984 & \underline{0.994} $\rightarrow{}$ 0.992 & \underline{0.993} $\rightarrow{}$ \underline{0.993} & \underline{0.869} $\rightarrow{}$ 0.834 & \underline{0.996} $\rightarrow{}$ 0.964 & \underline{0.992} $\rightarrow{}$ 0.978 &  \textbf{0.997} $\rightarrow{}$ 0.993 \\
                Tile  & 0.562 $\rightarrow{}$ \underline{0.697} & 0.836 $\rightarrow{}$ \underline{0.911} & 0.658 $\rightarrow{}$ \underline{0.670} & 0.576 $\rightarrow{}$ \underline{0.651} & \underline{0.811} $\rightarrow{}$ 0.802 & \underline{0.712} $\rightarrow{}$ 0.620 & 0.601 $\rightarrow{}$ \underline{0.609} & \underline{0.847} $\rightarrow{}$ 0.785 & \underline{0.744} $\rightarrow{}$ 0.714 &  0.847 $\rightarrow{}$ \textbf{0.911} \\
                Wood  & \underline{1.000} $\rightarrow{}$ 0.994 & 0.995 $\rightarrow{}$ \underline{1.000} & \underline{1.000} $\rightarrow{}$ 0.997 & 0.988 $\rightarrow{}$ \underline{0.999} & \underline{0.994} $\rightarrow{}$ 0.992 & 0.991 $\rightarrow{}$ \underline{0.995} & 0.987 $\rightarrow{}$ \underline{0.999} & 0.996 $\rightarrow{}$ \underline{0.999} & \underline{0.999} $\rightarrow{}$ 0.997 &  \textbf{1.000} $\rightarrow{}$ \textbf{1.000} \\
            \hline
            \hline
                \textbf{Average}  & 0.840 $\rightarrow{}$ \underline{0.851} & 0.885 $\rightarrow{}$ \underline{0.891} & \underline{0.874} $\rightarrow{}$ 0.861 & 0.837 $\rightarrow{}$ \underline{0.848} & \underline{0.878} $\rightarrow{}$ 0.877 & 0.860 $\rightarrow{}$ \underline{0.864} & 0.798 $\rightarrow{}$ \underline{0.799} & \underline{0.874} $\rightarrow{}$ 0.863 & \underline{0.863} $\rightarrow{}$ \underline{0.863} &  0.908 $\rightarrow{}$ \textbf{0.913} \\
        \end{tabular}
    }
    \label{table:max_auroc}
    \vspace*{-0.3cm}
\end{table*}

\subsection{Comparison of loss landscapes}
\label{subsec:ladscape} 

We visualize the loss landscape of $\mathcal{L}_{2}$ and $\mathcal{L}_{2}^{LAMP}$ for the batch sizes of 128, 16, and 4.
$\mathcal{L}_{2}^{LAMP}$ denotes $\mathcal{L}_{2}$ metric amplified by LAMP.
For visualization, an open source\footnote{https://github.com/tomgoldstein/loss-landscape} provided by Li et al.~\cite{Landscape_Li_NeurIPS18} is used. 
The visualization results in Fig.~\ref{fig:loss_contour} show the loss landscape only for the encoder part of the AE that is trained to predict class labels of the MNIST dataset~\cite{MNIST_Yan_PIEEE98}.
The results indicate that all $\mathcal{L}_{2}^{LAMP}$ cases always show dense contours which means steeper loss landscapes compared to $\mathcal{L}_{2}$.

\begin{figure}[t]
    \setlength{\tabcolsep}{0pt}
    \resizebox{\columnwidth}{!}{%
        \begin{tabular}{ccc|ccc}
            \multicolumn{3}{c|}{\textbf{Normal}} & \multicolumn{3}{c}{\textbf{Defective}} \\
            % \hline
            \textbf{Input} & $\mathcal{L}_{base}$ & $\mathcal{L}_{base}^{LAMP}$ & \textbf{Input} & $\mathcal{L}_{base}$ & $\mathcal{L}_{base}^{LAMP}$ \\
            
            \includegraphics*[width=0.25\columnwidth]{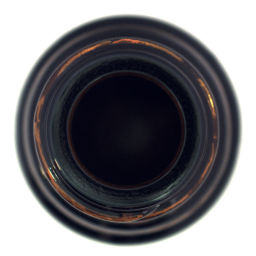} &
            \includegraphics*[width=0.25\columnwidth]{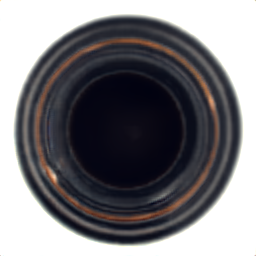} &
            \includegraphics*[width=0.25\columnwidth]{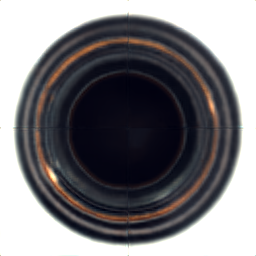} &
            
            \includegraphics*[width=0.25\columnwidth]{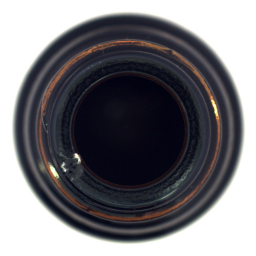} &
            \includegraphics*[width=0.25\columnwidth]{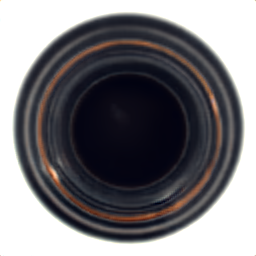} &
            \includegraphics*[width=0.25\columnwidth]{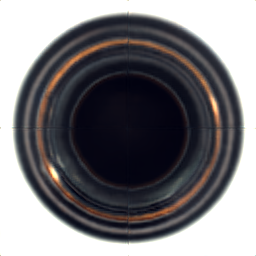} \\ 
            \vspace*{-0.55cm} \\
            
            \includegraphics*[width=0.25\columnwidth]{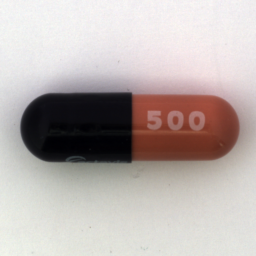} &
            \includegraphics*[width=0.25\columnwidth]{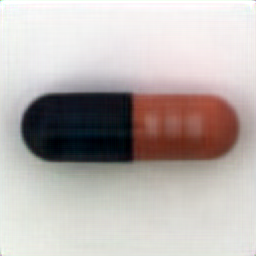} &
            \includegraphics*[width=0.25\columnwidth]{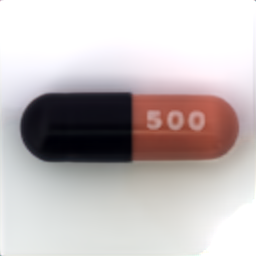} &
            
            \includegraphics*[width=0.25\columnwidth]{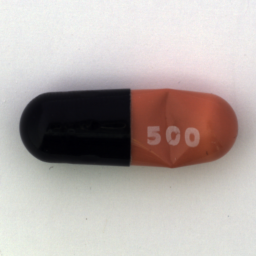} &
            \includegraphics*[width=0.25\columnwidth]{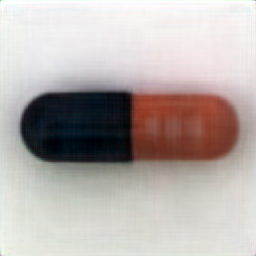} &
            \includegraphics*[width=0.25\columnwidth]{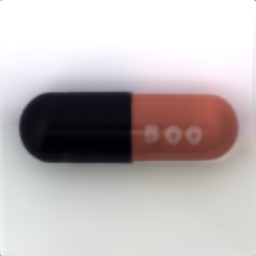} \\
            \vspace*{-0.55cm} \\
            
            \includegraphics*[width=0.25\columnwidth]{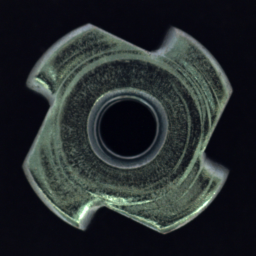} &
            \includegraphics*[width=0.25\columnwidth]{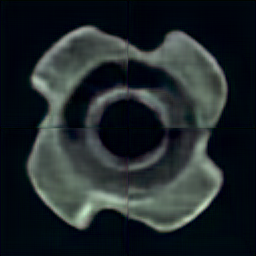} &
            \includegraphics*[width=0.25\columnwidth]{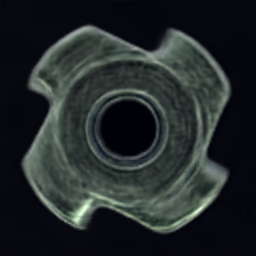} &
            
            \includegraphics*[width=0.25\columnwidth]{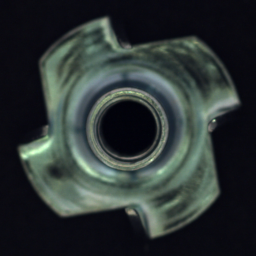} &
            \includegraphics*[width=0.25\columnwidth]{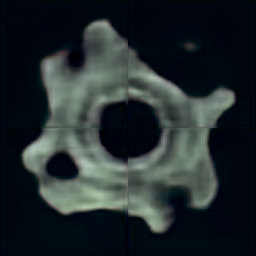} &
            \includegraphics*[width=0.25\columnwidth]{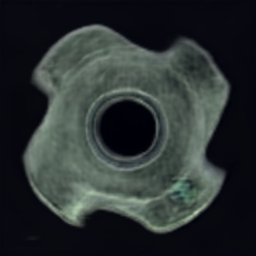} \\
            \vspace*{-0.55cm} \\
            
            \includegraphics*[width=0.25\columnwidth]{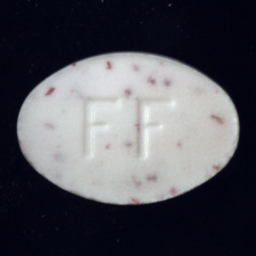} &
            \includegraphics*[width=0.25\columnwidth]{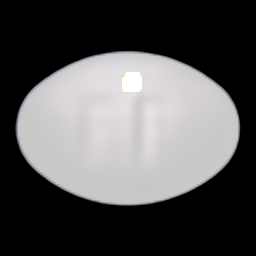} &
            \includegraphics*[width=0.25\columnwidth]{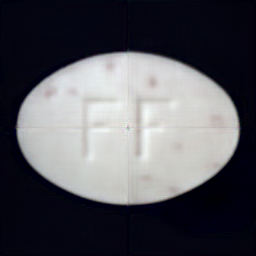} &
            
            \includegraphics*[width=0.25\columnwidth]{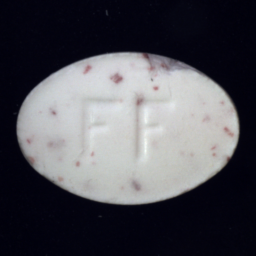} &
            \includegraphics*[width=0.25\columnwidth]{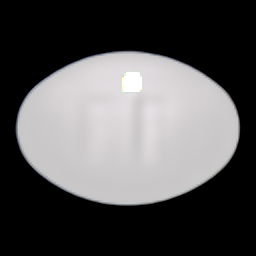} &
            \includegraphics*[width=0.25\columnwidth]{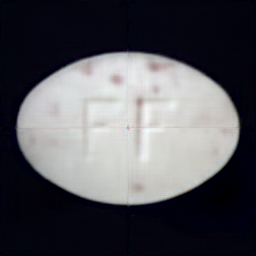} \\
            \vspace*{-0.55cm} \\
            
            \includegraphics*[width=0.25\columnwidth]{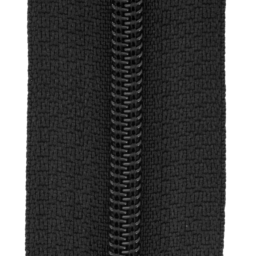} &
            \includegraphics*[width=0.25\columnwidth]{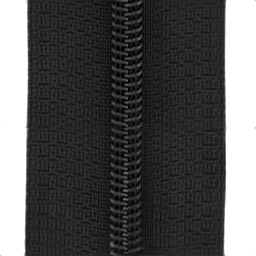} &
            \includegraphics*[width=0.25\columnwidth]{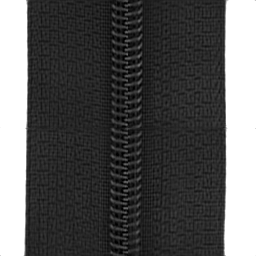} &
            
            \includegraphics*[width=0.25\columnwidth]{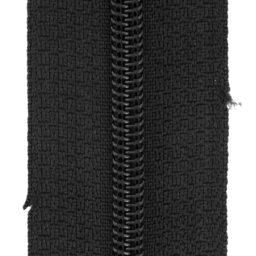} &
            \includegraphics*[width=0.25\columnwidth]{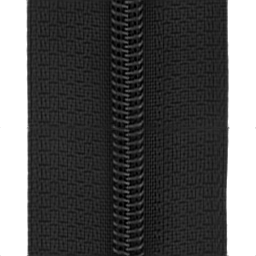} &
            \includegraphics*[width=0.25\columnwidth]{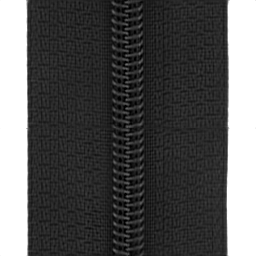} \\
            \vspace*{-0.8cm} \\
        \end{tabular}
    }
    \caption{Examples of reconstruction results. The $\mathcal{L}_{base}^{LAMP}$ case demonstrates improved reconstructions. Note the clear visibility of the number `500' on the normal capsule, which appears indistinct in $\mathcal{L}_{base}$. Also, $\mathcal{L}_{base}^{LAMP}$ cases transform the input data closer to the learned normal pattern, resulting in a more robust AD than $\mathcal{L}_{base}$.}
    \label{fig:generations}
    \vspace*{-0.2cm}
\end{figure} 

\subsection{AD experiments using MNIST dataset}
\label{subsec:toy}

Our work is based on the fact that when the loss landscape of an NN is shaped steeper, it will show better AD performance due to the contained generalization ability~\cite{Landscape_Li_NeurIPS18}.
The results shown in Table~\ref{table:mnist_batch} experimentally prove our claim that LAMP-applied cases transform the loss landscape steeper and always outperform regardless of the batch size.
Moreover, in a large batch size which yields a smoother loss landscape, the gap between AUROCs is greater, that is, the effect of LAMP is maximized.
This result experimentally demonstrates that the sharpness of the loss landscape can be exploited to enhance the AD performance.

\subsection{Results for industrial dataset}
\label{subsec:industrial}

We fix the batch size to 16 in this experiment considering the training speed.
Table~\ref{table:max_auroc} shows the experimental results on the industrial dataset, MVTec AD~\cite{MVTecAD_Paul_CVPR19}, in nine experimental settings from three different base loss functions and three optimizers.
In 5 out of 9 experimental settings, LAMP-applied cases achieve equal or better performances.
The last column of Table~\ref{table:max_auroc} shows the best performance for each subtask and $\mathcal{L}_{base}^{LAMP}$ attains better AUROC than $\mathcal{L}_{base}$. 
% At this moment, LAMP outperforms in 9 subtasks except for 2 tasks that are tied among 15 subtasks.

In Fig.~\ref{fig:generations}, we present reconstruction results for the best models trained with each $\mathcal{L}_{base}$ and $\mathcal{L}_{base}^{LAMP}$.
$\mathcal{L}_{base}$ produces blurry results for normal products in capsule, metal nut, and pill cases, causing a large reconstruction error.
Furthermore, a pill case shows fixed reconstruction results regardless of changing the input, undermining the reliability of the AD performance.
In contrast, $\mathcal{L}_{base}^{LAMP}$ case demonstrates accurate reconstructions for normal samples.
Note the clear visibility of the number `500' on the normal capsule.
In cases of unseen anomalous patterns, they are converted closer to a seen normal pattern, which is intended of a reconstruction model in UAD settings.
This yields accurate detection of defective samples.
Via the results of extensive experiments, we confirm that LAMP improves the AD performance in most cases quantitatively and qualitatively.

\section{Conclusion}
\label{sec:conclusion}

In this paper, we propose a simple method to enhance the AD performance in an UAD setting from the perspective of reconstruction loss amplification by noting that contained generalization ability is highly related to sharp-shaped loss landscapes.

To show the legitimacy of our approach, we design extensive experiments with MNIST and MVTec AD datasets.
We have shown the shape change of the reconstruction loss landscape when LAMP is applied as we vary the batch size.
We demonstrate quantitative and qualitative performance enhancement of an UAD model by LAMP under various conditions.
LAMP can be safely applied to any reconstruction error metrics in an UAD setup where a reconstruction model is trained with anomaly-free samples only.

% \pagebreak
\balance
\bibliographystyle{IEEEbib}
% \bibliography{strings,refs}

\end{document}